\documentclass[11pt]{article}

\usepackage[margin=1in]{geometry}
\usepackage[T1]{fontenc}
\usepackage[utf8]{inputenc}
\usepackage{times}
\usepackage{microtype}
\usepackage{amsmath,amssymb}
\usepackage{graphicx}
\usepackage{booktabs}
\usepackage{array}
\usepackage{tabularx}
\usepackage{multirow}
\usepackage{longtable}
\usepackage{enumitem}
\usepackage{hyperref}
\usepackage{natbib}
\usepackage{url}
\usepackage{tikz}
\usetikzlibrary{arrows.meta,positioning}
\usepackage{caption}
\usepackage{xcolor}
\usepackage{mdframed}

\hypersetup{colorlinks=true,linkcolor=blue,citecolor=blue,urlcolor=blue,
	pdftitle={Constraint-Driven Model Optimization}}

\definecolor{boxgray}{RGB}{242,242,242}
\newmdenv[backgroundcolor=boxgray,linewidth=0pt,
innerleftmargin=8pt,innerrightmargin=8pt,
innertopmargin=6pt,innerbottommargin=6pt]{takeawaybox}

\tikzset{
	pnode/.style={rectangle,rounded corners=3pt,draw=black,fill=gray!12,
		text width=6cm,align=center,minimum height=0.65cm,font=\small},
	parr/.style={-Stealth,semithick}
}

\begin{document}
	
	\title{\bf Constraint-Driven Model Optimization:\\ An Industry Framework
		for Selecting Compression and Acceleration Techniques\\
		in Modern Machine Learning Systems}
	
	\author{
		\textbf{Dhruv Shivkant}\thanks{Author order is based on alphabetical ordering.}\thanks{Work done during internship at EXL.}\\
		Indian Institute of Science\\
		\texttt{dhruvsp@iisc.ac.in}
		\and
		\textbf{Saket Mohanty}\\
		EXL\\
		\texttt{Saket.Mohanty@exlservice.com}
		\and		
        \textbf{Somya Rai}\\
		EXL\\
		\texttt{Somya.Rai@exlservice.com} 
        \and
		\textbf{Utkarsh Wadhwa}\\
		EXL\\
		\texttt{Utkarsh.Wadhwa@exlservice.com}
	}
	\date{}
	\maketitle
	\thispagestyle{empty}
	
	\begin{abstract}
		The rapid deployment of machine learning systems across cloud, edge, and
		enterprise environments has brought model optimization to the forefront of systems-engineering. Despite a rich literature
		spanning quantization, pruning, knowledge distillation, parameter-efficient
		fine-tuning (PEFT), and inference-time optimization, practitioners are often
		left navigating these techniques through heuristics rather than principled
		methodology. We argue that optimization should be formulated as a
		\emph{constraint-driven, multi-objective engineering decision} and introduce
		a unified framework that characterizes any production deployment along five
		interacting constraint dimensions: \textbf{data availability}, \textbf{latency budget},
		\textbf{memory budget}, \textbf{accuracy tolerance}, and \textbf{retraining
			budget}. Building on this taxonomy, we synthesize empirical gains reported
		across the research literature 
		and map them to operational
		constraints rather than algorithmic categories. To ensure practical relevance, we selected these techniques by reviewing recent literature for methods that report measurable improvements against critical deployment bottlenecks. We propose a prescriptive
		decision framework and provide optimization pipelines for four representative
		industrial scenarios to illustrate it in practice. To the best of our knowledge, this work provides one of the first structured attempts to
		formalize model optimization as a constraint-aware, multi-objective engineering
		process, synthesizing quantitative evidence from the research literature.
	\end{abstract}
	\newpage
	\tableofcontents
	\newpage
	
	\section{Introduction}
	
	\subsection{The Deployment Imperative}
	
	Modern machine learning's research interest, particularly in Large Language Models (LLMs), is shifting from furthering model capabilities toward serving them efficiently at scale. LLMs introduce challenges that go beyond raw parameter count: autoregressive decoding is sequential and does not fully exploit the parallel processing capabilities of GPUs, the Key-Value (KV) cache grows proportionally with sequence length, number of layers, and attention heads (making its total footprint substantial for large models under long-context or high-concurrency workloads), and deployment targets span a wide range of resource constraints from data-center GPUs to embedded microcontrollers. Practical use often dictates that reasonable accuracy at lower latency and computational cost is more valuable than a state-of-the-art model that demands heavy compute or incurs high latency. For many production systems, inference has consequently become the dominant ongoing cost center, often exceeding the amortized cost of training over a model's operational lifetime.
	
	
	\subsection{A Practitioner's Gap}
	
	Despite an extensive research literature, a systematic gap persists between how
	optimization techniques are studied and how they are deployed in practice.
	Academic work frames optimization as an algorithmic question:
	\begin{quote}
		\emph{``How can technique X be improved on benchmark Y?''}
	\end{quote}
	By contrast practitioners face a different, operationally grounded question:
	\begin{quote}
		\emph{``Given my hardware, latency SLA, data budget, and accuracy floor,
			which combination of techniques should I apply, and in what order?''}
	\end{quote}
	
	This paper directly addresses the second question. We do not introduce a new
	optimization algorithm. Instead we provide a structured decision framework that
	maps deployment constraints to optimization strategies, synthesizing quantitative
	evidence from the primary literature to support each recommendation. We reviewed recent proceedings from major machine learning and systems venues, selecting techniques that report quantifiable improvements against critical deployment bottlenecks such as latency, memory footprint, or retraining cost. We note that our review follows a narrative rather than a formal systematic-review methodology and does not provide a reproducible search protocol, inclusion/exclusion criteria, or exhaustive screening workflow.
	
	\subsection{Contributions}
	
	\begin{enumerate}[leftmargin=*, label=\textbf{\arabic*.}, noitemsep]
		\item \textbf{Operational Constraint Taxonomy.} Five interacting deployment constraint dimensions
		that collectively characterize a production ML system and provide a common
		vocabulary for optimization decisions.
		
		\item \textbf{Functional Categorization.} We note the functional category of each optimization technique (e.g., parameter reduction, inference optimization) alongside our primary constraint-based classification, complementing existing algorithm-centric surveys.
		
		\item \textbf{Quantitative Constraint--Technique Mapping.} A synthesis of
		empirical results from over 25 primary papers, mapping measured gains to the
		deployment constraints they address.
		
		\item \textbf{Four Prescriptive Deployment Pipelines.} Evidence-informed,
		step-by-step optimization pipelines for representative industrial deployment scenarios,
		with explicit constraint profiles and expected outcomes.
	\end{enumerate}


	\section{Operational Constraint Taxonomy}
	
	We characterize any machine learning deployment environment along five interacting constraint dimensions. While presented separately, these dimensions are not independent. For example, quantization simultaneously affects memory, latency, and accuracy, while data availability directly constrains the retraining budget. Together they define a constraint profile that can guide which optimization strategies to consider and in what sequence.
	
	\subsection{Data Availability}
	Data availability refers to the quantity, quality, and accessibility of labeled, unlabeled, or weakly labeled data usable for training, calibration, or adaptation. In many enterprise settings, teams operate with zero task-specific data, meaning only the pretrained checkpoint is accessible. This restricts optimization to training-free techniques like post-training quantization. In other scenarios, teams might possess limited data, consisting of less than 1,000 labeled examples or weak supervision signals, which opens the door to parameter-efficient fine-tuning (though with such small datasets, full PEFT on large models still carries a risk of overfitting). When moderate data of 1,000 to 100,000 labeled examples is available, full alignment tuning becomes viable. Finally, abundant data involving hundreds of thousands of examples permits full retraining and large-scale knowledge distillation.
	
	\subsection{Latency Budget}
	The latency budget defines the maximum acceptable response time, characterized by the target latency percentile and the specific deployment mode. The distinction between time-to-first-token, inter-token latency, and end-to-end latency matters and should be specified explicitly. Real-time applications, such as autonomous systems or live voice agents, demand strict latency under 200 milliseconds, typically requiring techniques such as speculative decoding, early exiting, or model distillation to a smaller architecture. Interactive applications, typified by chat interfaces or code completion assistants, generally target between 200 milliseconds and 2 seconds. Batch or offline processing pipelines prioritize raw throughput and hardware utilization over per-query latency, making continuous batching particularly effective in this regime.
	
	\subsection{Memory Budget}
	Memory budget encapsulates the total available memory spanning both Video RAM (VRAM) and standard DRAM. This budget must accommodate parameter storage, the key-value (KV) cache, runtime activations, and optimizer states if training is involved. Edge and mobile deployments represent the most constrained environments, typically offering less than 4 GB of VRAM. Consumer GPUs provide a moderate budget of up to 24 GB of VRAM, allowing for local execution of quantized multibillion-parameter models. Enterprise GPU nodes, often spanning 80 GB to 320 GB of VRAM, shift the memory bottleneck away from parameter storage and heavily toward managing the KV cache for thousands of concurrent user sessions.
	
	\subsection{Accuracy Tolerance}
	Accuracy tolerance dictates the maximum permissible performance degradation following optimization relative to the original, unoptimized reference model. The specific metric must be defined per-deployment (for example, exact match, task success rate, factuality, calibration error, or human preference win rate). The thresholds below are illustrative and not directly comparable across metrics. Mission-critical deployments, spanning medical diagnostics or financial trading, typically permit less than 0.5\% degradation on the chosen task metric and may restrict optimization to well-calibrated 8-bit quantization. Business-critical applications, encompassing standard customer support or internal productivity tools, generally tolerate between 0.5\% and 2\% degradation, allowing for 4-bit quantization and structured pruning. Cost-sensitive applications, where high query volume makes serving the full model financially unsustainable, may accept greater than 2\% degradation in exchange for substantial throughput gains. In all cases, practitioners should also evaluate tail-risk failures, subgroup performance, and severity-weighted errors rather than relying solely on aggregate accuracy.
	
	\subsection{Retraining Budget}
	The retraining budget quantifies the available GPU-hours, engineering time, and financial resources allocated for modifying the deployed model. A retraining budget of zero means no compute is available to alter model weights, which limits practitioners to post-training quantization, training-free sparsification, and inference engine optimizations. A minimal budget, typically under 10 GPU-hours on a single node, can support targeted parameter-efficient fine-tuning strategies. A full budget representing hundreds of GPU-hours on cluster hardware can justify more extensive procedures such as quantization-aware training, neural architecture search, and structured pruning.
	
	\subsection{Scope and Limitations of This Taxonomy}
	The five dimensions above provide a useful initial characterization, but production deployments involve additional first-class constraints that this framework does not treat as primary axes, including:
	\begin{itemize}
		\item Throughput and goodput
		\item Concurrency, traffic shape, and burstiness
		\item Prompt and output-length distributions
		\item Cost per successful request (serving cost)
		\item Energy and thermal limits
		\item Hardware and kernel compatibility
		\item Availability, resiliency, and observability
		\item Data privacy and model residency
	\end{itemize}
	Furthermore, latency and throughput should not be treated as interchangeable, as optimizing one can hurt the other. We encourage practitioners to extend the constraint profile to cover these dimensions as appropriate for their deployment.
	
	\section{Optimization Techniques by Constraint Dimension}
	
	Rather than grouping by algorithmic family, we organize the literature by the primary deployment constraint each technique addresses. This provides practitioners with a direct mapping from their binding constraint to candidate solutions. (An overview of all evaluated techniques is provided in Table~\ref{tab:comparison} in the Appendix.)
	
	\subsection{Memory Budget Optimization}
	Techniques in this category reduce the memory footprint of the model parameters or the runtime KV cache, enabling deployment on smaller or fewer GPUs.
	
	\textbf{GPTQ \citep{frantar2023gptq}:} A post-training weight quantization technique that uses an approximate inverse-Hessian method to minimize layer-wise reconstruction error. By quantizing billion-scale models to 3 or 4 bits, it achieves approximately a 4$\times$ reduction in weight-storage memory footprint relative to FP16. In the reported experiments, the resulting 4-bit models maintain a perplexity within 0.25 points of the FP16 baseline, and the quantization process required less than 4 GPU-hours for a 175B model on the hardware used in the original study. These figures will vary with model architecture, hardware, and configuration.
	
	\textbf{AWQ \citep{lin2024awq}:} An activation-aware weight quantization technique that protects a small subset (roughly 1\%) of salient weight channels identified via activation magnitudes, avoiding the need for backpropagation during quantization. In the reported experiments, AWQ delivered accuracy within 0.5\% of the FP16 baseline on instruction-following benchmarks and provided a 1.45$\times$ throughput advantage over GPTQ on the tested edge hardware. Actual accuracy and throughput will depend on the model, task, and target device.
	
	\textbf{OmniQuant / ZeroQuant-V2 \citep{shao2023omniquant,yao2023zeroquant}:} Advanced quantization strategies designed to handle significant outliers in LLM activations. OmniQuant optimizes weight clipping boundaries and scaling factors, while ZeroQuant-V2 utilizes Low-Rank Compensation (LoRC) to mitigate quantization inaccuracies. These methods report near-lossless 4-bit and 8-bit quantization on the evaluated benchmarks without extensive retraining, though quality preservation will vary by model and task.
	
	\textbf{KIVI / KVQuant \citep{liu2024kivi,hooper2024kvquant}:} Specialized quantization algorithms targeting the Key-Value (KV) cache, which becomes the primary memory bottleneck in long-context inference. KIVI introduces a tuning-free asymmetric 2-bit quantization scheme for the KV cache. In the reported experiments, this approach substantially reduces runtime memory consumption, with results suggesting up to 4$\times$ longer effective context lengths on identical hardware with limited perplexity degradation. Gains are model- and sequence-length-dependent.
	
	\textbf{Wanda \citep{sun2023wanda}:} A training-free unstructured pruning approach that determines weight importance by evaluating the product of weight magnitudes and the norm of corresponding input activations. Wanda achieves 50\% unstructured sparsity in LLMs while often outperforming standard magnitude pruning by several percentage points on zero-shot benchmarks. (It is included in this category because it targets parameter compression, though setting weights to zero does not automatically reduce physical memory, deployed model size, or inference latency. Realizing these runtime benefits requires sparse storage formats, compatible sparse kernels, and hardware or runtime support for the chosen sparsity pattern).
	
	\textbf{FlexGen / PowerInfer \citep{sheng2023flexgen,song2023powerinfer}:} System-level offloading frameworks designed to execute large LLMs on hardware with limited GPU VRAM. FlexGen offloads weights and KV cache to CPU memory and disk, formulating a graph traversal problem to overlap data movement with computation. PowerInfer exploits activation sparsity to keep hot neurons on the GPU and cold neurons on the CPU. In the reported experiments these methods achieve substantially higher single-GPU generation throughput compared to naive offloading (up to 10$\times$ in some configurations), though actual gains depend heavily on I/O bandwidth, sparsity pattern, and model architecture.
	
	\textbf{LoSparse \citep{li2023losparse}:} A hybrid approach combining low-rank approximation with weight pruning. By leveraging low-rank decomposition, LoSparse prevents the loss of expressive neurons that typically occurs with direct model pruning. This strategy can provide significant parameter reduction while better maintaining the reasoning capabilities of the original model compared to pruning alone, though the actual memory and latency benefits depend on the availability of compatible sparse kernels and runtime support.
	
	\subsection{Latency Budget Optimization}
	These methods aim to accelerate inference speed or increase throughput, helping systems work toward real-time or interactive latency service level agreements. SLA compliance also depends on hardware configuration, serving infrastructure, and workload characteristics.
	
	\textbf{FlashAttention \citep{dao2022flashattention} / FlashAttention-2 \citep{dao2023flashattention2fasterattentionbetter}:} IO-aware exact attention algorithms that fuse softmax and matrix multiplications into tiled CUDA kernels to minimize High Bandwidth Memory (HBM) accesses. FlashAttention (2022) reduces auxiliary attention-memory requirements from quadratic to linear in sequence length (the attention computation itself remains $O(N^2)$ for dense full attention), reporting 2--4$\times$ wall-clock speedups over optimized baselines in the original experiments. FlashAttention-2 (2023) further improves GPU work partitioning and parallelism, achieving approximately 2$\times$ additional speedup over FlashAttention in the reported benchmarks. Speedup figures are workload-, hardware-, and baseline-dependent.
	
	\textbf{Speculative Decoding (Medusa / Eagle) \citep{cai2024medusa,li2023eagle}:} Advanced decoding strategies that break the sequential bottleneck of autoregressive generation. While traditional speculative decoding uses a separate draft model, Medusa trains multiple heads on top of the target model to predict subsequent tokens, and Eagle predicts the target model's next hidden state to generate draft tokens without a separate model, reducing draft generation cost. These methods achieve 2 to 3.7$\times$ end-to-end wall-clock speedups in the reported experiments. When the verification and rejection-sampling procedure is exact, speculative sampling can preserve the target model's output distribution, though bit-for-bit identical outputs are not guaranteed across all implementations, RNG configurations, and acceptance strategies. Performance benefits are also workload-dependent, as speculative decoding can regress throughput under high concurrency, poor draft-acceptance rates, or memory-constrained settings.
	
	\textbf{vLLM (PagedAttention) \citep{kwon2023vllm}:} A serving engine optimization that manages the KV cache using principles inspired by operating system virtual memory. By storing the KV cache in non-contiguous paged blocks, PagedAttention substantially reduces memory fragmentation, though block-level internal fragmentation and scheduling overhead can remain. This continuous batching strategy allows serving systems to process more concurrent requests (up to 4$\times$ in the reported experiments), improving the throughput capabilities of GPU clusters.
	
	\textbf{LLMLingua \citep{jiang2023llmlingua}:} A prompt compression technique that performs coarse-to-fine pruning of input tokens based on perplexity signals from a smaller model. By removing redundant tokens before the prefilling stage, LLMLingua reduces the initial computation time. In the reported experiments it compresses prompts by up to 20$\times$ while preserving downstream task quality. The degree of acceptable compression and quality retention is task- and model-dependent.
	
	\textbf{Skeleton-of-Thought (SoT) \citep{ning2023skeleton}:} An output organization framework that instructs the LLM to first generate a concise skeleton of points, and then uses batch inference to expand upon each point in parallel. In the reported experiments, SoT reduces end-to-end generation latency by up to 2.39$\times$ on multi-part queries without modifying the underlying model architecture. Gains depend on query structure and the availability of parallel decoding capacity (which trades per-query latency for overall system throughput).
	
	\textbf{StreamingLLM \citep{xiao2023streamingllm}:} A framework that enables LLMs trained with finite context windows to maintain stable generation over very long token streams without fine-tuning. By recognizing that LLMs rely heavily on initial tokens (attention sinks), StreamingLLM preserves these sink tokens alongside a sliding window of recent tokens and discards intermediate KV states. This permits stable language modeling over millions of processed tokens with bounded memory usage. Because intermediate tokens are evicted, however, StreamingLLM does not provide faithful semantic access to all earlier tokens. It is suited for streaming or dialogue continuity, not for tasks requiring recall or reasoning over the full document history.

	\textbf{Gisting / AutoCompressors \citep{mu2023learning,chevalier2023adapting}:} Soft prompt compression techniques that condense task-specific instructions into a small set of learnable continuous tokens. Gisting uses prefix-tuning to create virtual tokens that summarize prompts, while AutoCompressors utilize an autoencoder approach over the context. These methods can substantially reduce the token footprint of task instructions for downstream use (reducing prefill latency), though compression quality and resulting task performance depend on the model and instruction type.
	
	\subsection{Data Availability Optimization}
	These strategies optimize model performance when labeled domain data is highly scarce or non-existent, leveraging synthetic data or parameter-efficient prompting.
	
	\textbf{LIMA \citep{zhou2023lima}:} A study suggesting that alignment and instruction-following capabilities of an LLM may be strongly influenced by data quality rather than quantity alone. By fine-tuning a 65B parameter model on a carefully curated dataset of only 1,000 examples, the resulting model was preferred over GPT-4 in 43\% of human evaluations in the reported study. The authors argue this reduces reliance on large-scale labeled datasets or Reinforcement Learning from Human Feedback (RLHF), though the generalizability of this finding to other model families and tasks requires further investigation.
	
	\textbf{Socratic CoT \citep{shridhar2023distilling}:} A knowledge distillation approach that addresses the lack of high-quality reasoning data. It fine-tunes a pair of models (a Question Generation model and a Question Answering model) to synthesize intermediate reasoning steps, then distills these synthetic rationales from a large teacher LLM into a smaller student. The approach aims to transfer Chain-of-Thought reasoning capabilities without requiring human-annotated reasoning paths. Reported gains are benchmark- and domain-specific.
	
	\subsection{Retraining Budget Optimization}
	These methods reduce the compute and data requirements for model adaptation. It is worth noting that Parameter-Efficient Fine-Tuning (PEFT) techniques such as LoRA primarily reduce \emph{training} resource requirements and do not automatically improve inference latency. Merged adapters match the base model's inference cost, while unmerged adapters add overhead. Similarly, Mixture-of-Experts (MoE) architectures are model-design choices and are not post-hoc optimization steps that can be applied to an existing dense checkpoint.
	
	\textbf{LoRA \citep{hu2022lora}:} Low-Rank Adaptation freezes the pretrained model weights and injects trainable rank decomposition matrices into the architecture. When applied to a 175B parameter model in the original experiments, LoRA reduced the number of trainable parameters by approximately 10,000$\times$ to 0.01\% of total parameters, achieving comparable performance to full fine-tuning on benchmarks such as SuperGLUE while reducing GPU memory requirements for the training step (primarily by reducing optimizer state overhead) by a reported factor of 3. These savings apply to the adaptation phase; at inference, merged adapters incur no additional cost over the base model.
	
	\textbf{LoftQ / QA-LoRA \citep{li2023loftq,xu2023qalora}:} Techniques that bridge the gap between post-training quantization and quantization-aware training. LoftQ initializes the LoRA matrices using Singular Value Decomposition (SVD) of the difference between the original weights and the quantized weights. QA-LoRA utilizes group-wise quantization and merges LoRA terms into the quantized matrices. Both methods are designed to enable fine-tuning of heavily compressed models on consumer hardware with reduced accuracy loss compared to naive quantization followed by LoRA. Actual fidelity depends on quantization level and task.

	\textbf{LLM-QAT / Norm Tweaking \citep{liu2023llmqat,li2023norm}:} Quantization-Aware Training methods. LLM-QAT generates its own synthetic training data by prompting the FP16 model, and then uses a distillation workflow to train the quantized model. Norm Tweaking fine-tunes only the LayerNorm layers using generated data, aiming for QAT-level accuracy recovery with lower data and compute requirements than full QAT. Actual savings are model- and task-dependent. (While they avoid requiring the original training data, they do require retraining compute budgets).
	
	\textbf{Switch Transformers / Expert Choice \citep{fedus2022switch,zhou2022mixture}:} Efficient structural designs utilizing Mixture-of-Experts (MoE) to scale model capacity without proportional increases in training compute. Switch Transformers route tokens to a single expert, minimizing routing overhead. Expert Choice allows experts to select the top-$k$ tokens, which tends to improve load balancing compared to token-choice routing, though perfect balancing is not guaranteed in all configurations. These MoE architectures achieve better task performance for the same training compute budget compared to dense models of similar active parameter count and offer efficient fine-tuning pathways, but adopting them requires selecting or training an MoE model from scratch rather than converting an existing dense checkpoint.
	
	\textbf{LoraPrune / ZipLM \citep{zhang2023loraprune,kurtic2023ziplm}:} Methods that combine structured pruning with parameter-efficient fine-tuning. LoraPrune uses the weights and gradients of LoRA matrices as a criterion to iteratively prune unimportant weights. ZipLM iteratively identifies and removes structural components offering the worst loss-to-runtime tradeoff. In the reported experiments, these methods produce compressed, domain-adapted models with comparatively low GPU-hour budgets. Actual compute requirements depend on model size, pruning target, and task.
	
	\subsection{Validate Accuracy and Manage Cost}
	Techniques for managing financial costs and maintaining reliability, factuality, and calibration post-optimization.
	
	\textbf{FrugalGPT \citep{chen2023frugalgpt}:} A cost-reduction strategy employing a cascade router that directs queries to APIs of varying capability based on a learned complexity threshold. In the original study's workload, approximately 83\% of queries were handled by cheaper, smaller models, with the remaining 17\% routed to the premium API, yielding up to 98\% cost reduction with a 4\% accuracy improvement (though this accuracy improvement is counterintuitive and likely due to ensemble effects on those specific benchmarks; generally, routing more queries to smaller models trades accuracy for cost). These routing ratios and cost savings are workload-specific and should not be treated as universal constants. Practitioners must measure escalation rates on their own traffic distribution.
	
	\textbf{SpQR / SqueezeLLM \citep{dettmers2023spqr,kim2023squeezellm}:} Accuracy-preserving quantization methods that identify a small fraction of weight outliers disproportionately affecting model accuracy. SpQR isolates and stores these outliers in FP16 format while heavily quantizing the remaining weights to 3 or 4 bits. This outlier-aware approach reports near-lossless compression on the evaluated benchmarks, with the aim of preserving reasoning and calibration quality. Performance preservation is model- and task-dependent.
	
	\textbf{CALM / SkipDecode \citep{schuster2022confident,delcorro2023skipdecode}:} Dynamic inference techniques that implement early exiting during the decoding phase. CALM trains classifiers after each layer to output confidence scores, halting inference early for tokens where confidence is high. SkipDecode extends this to batch inference by ensuring a unified exit point for all tokens in a batch. In the reported experiments, these methods reduce the average computational cost per token with limited accuracy degradation. The trade-off between compute savings and quality is threshold- and task-dependent.
	
	\section{Decision-Making Framework}
	
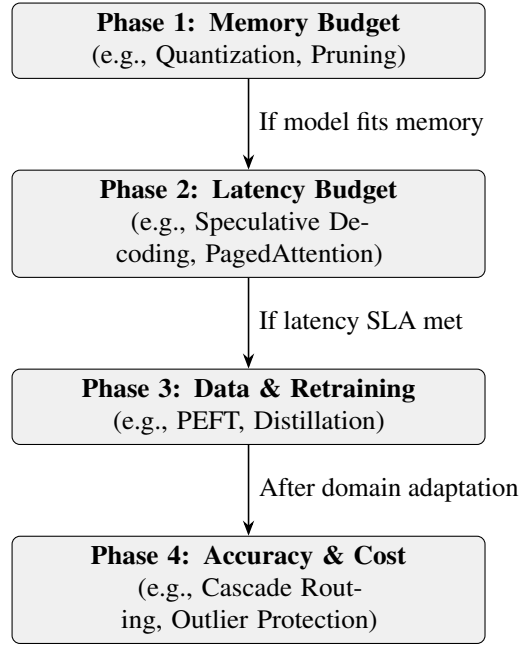
\begin{figure}[h!]
		\centering
		\begin{tikzpicture}[node distance=1.2cm, auto]
			\node[pnode] (mem) {\textbf{Phase 1: Memory Budget}\\(e.g., Quantization, Pruning)};
			\node[pnode] (lat) [below=of mem] {\textbf{Phase 2: Latency Budget}\\(e.g., Speculative Decoding, PagedAttention)};
			\node[pnode] (data) [below=of lat] {\textbf{Phase 3: Data \& Retraining}\\(e.g., PEFT, Distillation)};
			\node[pnode] (acc) [below=of data] {\textbf{Phase 4: Accuracy \& Cost}\\(e.g., Cascade Routing, Outlier Protection)};
			
			\draw[parr] (mem) -- node[right] {\small If model fits memory} (lat);
			\draw[parr] (lat) -- node[right] {\small If latency SLA met} (data);
			\draw[parr] (data) -- node[right] {\small After domain adaptation} (acc);
		\end{tikzpicture}
		\caption{Hierarchical Optimization Decision Framework. The phases are intended to be addressed in order, though practitioners may need to iterate across phases as constraints interact.}
		\label{fig:flowchart}
	\end{figure}

	To navigate the landscape of optimization techniques, we propose a structured, multi-phase methodology as a starting point. The framework recommends addressing optimization hierarchically: resolving hard infrastructural limits first, followed by sequential refinements to latency, domain adaptation, and accuracy assurance. This ordering is a heuristic guide and not a strict rule. Practitioners may need to revisit earlier phases as later constraints are evaluated. Figure~\ref{fig:flowchart} illustrates this progressive approach.
	
	\subsection{Phase 1: Determine Deployment Topology (Memory Budget)}
	Hardware memory limits are typically the first constraint to resolve, as they determine whether a model can be loaded at all.
	\begin{itemize}
		\item \textbf{Edge and Mobile Deployments ($<$8 GB VRAM):} In environments where parameter memory is the binding bottleneck, Post-Training Quantization (PTQ) is typically the first tool to reach for. Activation-aware methods like AWQ are well-suited to this regime and are designed to reduce quality loss on constrained hardware. If memory constraints persist after quantization, structured pruning or hardware-supported N:M sparsity patterns (which have compatible accelerated kernels) can further reduce the effective parameter count. Unstructured pruning (such as Wanda) sets weights to zero but does not reduce physical memory or latency without sparse storage formats, compatible kernels, and runtime support.
		\item \textbf{Single Enterprise Node (24--80 GB VRAM):} Deploying large models (e.g., 70B parameters) on a single GPU typically requires low-bit quantization methods, such as OmniQuant or ZeroQuant-V2. If increasing batch sizes begin to induce out-of-memory errors, system-level offloading frameworks like FlexGen or PowerInfer can be introduced to utilize CPU RAM and disk storage for cold activations.
		\item \textbf{Distributed GPU Cluster ($>$80 GB VRAM):} While parameter storage is less constrained across a cluster, the runtime memory demanded by the KV cache can rapidly become the primary bottleneck as concurrent users scale. In these scenarios, KV cache quantization techniques (such as KIVI or KVQuant) are worth implementing early to compress the runtime footprint.
	\end{itemize}
	
	\subsection{Phase 2: Establish the Latency and Throughput SLA (Latency Budget)}
	Once the model fits within the memory budget, the serving engine can be tuned to work toward specific Service Level Agreements (SLAs).
	\begin{itemize}
		\item \textbf{Real-Time Interactive Systems (e.g., Voice and Code Completion):} When token generation speed is paramount, speculative decoding architectures like Eagle or Medusa are worth evaluating. In the original experiments they achieved 2--3.7$\times$ speedups, though gains are workload-dependent. Pairing this with FlashAttention-2 can reduce HBM traffic during the prefill phase, which may lower prefill latency.
		\item \textbf{High-Concurrency API Serving:} If maximizing system throughput is the primary objective, deploying a serving engine equipped with PagedAttention (like vLLM) enables continuous batching and substantially reduces KV cache fragmentation. Additionally, implementing batch-aware early exiting (e.g., SkipDecode) can dynamically halt generation for highly confident tokens, thereby reducing the average computational cost per request.
		\item \textbf{Long-Context Applications (e.g., Document QA):} Prefilling large context windows is computationally expensive. Prompt compression techniques like LLMLingua can prune redundant tokens from the input before processing, reducing prefill cost. For streaming or dialogue-continuity use cases, StreamingLLM can preserve essential attention sinks to maintain stable generation over very long token streams. StreamingLLM evicts intermediate KV states, however, and is therefore unsuitable for tasks requiring faithful recall over the entire document. For such tasks, retrieval reranking, hierarchical summarization, or validated long-context models should be considered instead.
	\end{itemize}
	
	\subsection{Phase 3: Assess the Data and Retraining Reality (Data \& Retraining Budgets)}
	Domain adaptation strategy is shaped by the intersection of available data and computational resources. The guidance below describes typical patterns, though the best approach should be validated empirically.
	\begin{itemize}
		\item \textbf{Abundant Data ($>$10K samples) with High Compute:} This scenario supports full task-specific distillation. By training a compact student model to match the logits and hidden states of a larger teacher model, practitioners can achieve faster inference while targeting domain accuracy retention. Actual accuracy versus speed trade-offs depend on model capacity, data quality, and distillation configuration.
		\item \textbf{Moderate Data ($\sim$1K curated samples) with Low Compute (Single GPU):} When compute is limited, Parameter-Efficient Fine-Tuning is a practical path. Methods like QA-LoRA or LoftQ integrate low-rank adaptation directly into the quantized weights, enabling fine-tuning without fully dequantizing the model. Quality outcomes depend on the rank, quantization level, and task.
		\item \textbf{Zero Task Data:} In the complete absence of training data, traditional retraining is impossible. Practitioners must rely on quantization methods that do not require the original training data, such as LLM-QAT or Norm Tweaking. These techniques generate synthetic calibration data from the pretrained model itself, enabling quantization-aware training without human-labeled datasets (though they are not strictly ``zero-data'' in an operational sense, as they do require compute for synthetic data generation and calibration).
	\end{itemize}
	
	\subsection{Phase 4: Fortify Accuracy and Manage Cost (Accuracy Tolerance)}
	The final phase addresses accuracy assurance and cost management for the optimized system.
	\begin{itemize}
		\item \textbf{Mission-Critical Reliability (Medical, Legal, Finance):} Standard low-bit quantization can introduce tail-risk failures in highly sensitive domains, making quality preservation particularly important. Outlier-aware quantization (like SpQR or SqueezeLLM) isolates and stores high-magnitude weight outliers in FP16 while quantizing the remainder, and has been shown to better preserve model quality than uniform low-bit quantization. Self-Consistency sampling (generating multiple reasoning paths and taking a majority vote) can improve answer stability on reasoning tasks, but it does not guarantee factuality---correlated samples from the same model can repeat the same unsupported claims. For high-stakes domains, practitioners should additionally implement sentence-level citations, evidence-span validation, groundedness and entailment checks, answerability detection, and abstention thresholds.
		\item \textbf{API Cost Reduction:} When relying on proprietary external models, the optimization target shifts from GPU memory to financial expenditure. A cascade routing system such as FrugalGPT can reduce costs by directing simpler queries to a cheaper, locally hosted quantized model (e.g., an 8B parameter architecture) and reserving the premium API for complex reasoning tasks. As noted earlier, the cost savings are workload-specific and must be measured on the organization's own traffic distribution.
	\end{itemize}
	
	\section{Industry Use Cases}
	
	\subsection{Edge AI: Alice and the Mobile Deployment}
	Alice is a mobile AI engineer tasked with deploying a 7B parameter local assistant onto a smartphone with a strict 4 GB RAM limit and heavy battery constraints.

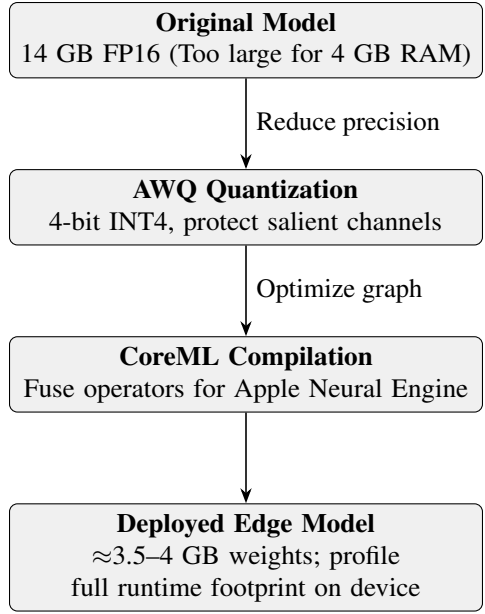
\begin{figure}[h!]
		\centering
		\begin{tikzpicture}[node distance=1.2cm, auto]
			\node[pnode] (start) {\textbf{Original Model}\\14 GB FP16 (Too large for 4 GB RAM)};
			\node[pnode] (awq) [below=of start] {\textbf{AWQ Quantization}\\4-bit INT4, protect salient channels};
			\node[pnode] (coreml) [below=of awq] {\textbf{CoreML Compilation}\\Fuse operators for Apple Neural Engine};
			\node[pnode] (end) [below=of coreml] {\textbf{Deployed Edge Model}\\$\approx$3.5--4 GB weights; profile full runtime footprint on device};
			
			\draw[parr] (start) -- node[right] {\small Reduce precision} (awq);
			\draw[parr] (awq) -- node[right] {\small Optimize graph} (coreml);
			\draw[parr] (coreml) -- (end);
		\end{tikzpicture}
		\caption{Optimization Pipeline for Edge AI Deployment. 4-bit weight quantization reduces the weight payload from 14 GB to approximately 3.5--4 GB; the total on-device footprint including runtime buffers and KV cache must be profiled separately.}
		\label{fig:usecase_edge}
	\end{figure}

	Facing a severe memory bottleneck, she begins by analyzing her toolkit. A standard 7B model at FP16 requires approximately 14 GB of memory for weights alone, which far exceeds the device budget. Alice's primary lever is 4-bit weight-only quantization: she applies AWQ, which identifies and protects salient weight channels while quantizing the remainder to INT4. A 7B model at 4-bit precision requires approximately 3.5--4 GB for weights (depending on the fraction of channels stored at higher precision and embedding layers), bringing it within the hardware limit. The total on-device footprint will also include runtime buffers, the KV cache (which scales with prompt length), tokenizer overhead, and the mobile runtime, so Alice must profile these components on the target device to confirm headroom. She then applies graph-level optimizations using CoreML to fuse operators for the device's Neural Engine, improving energy efficiency and token throughput. Should she also wish to apply Wanda pruning to further reduce the weight payload, she would need to verify that the target runtime supports sparse storage and sparse matrix operations. Without such support, unstructured sparsity provides no memory or latency benefit at inference time (see Figure~\ref{fig:usecase_edge}).

	\subsection{Enterprise LLM Serving: Bob and the GPU Cluster}
	Bob is the infrastructure lead for a high-traffic enterprise application, responsible for serving a 70B parameter open-weight model under heavy concurrent load.

\begin{figure}[h!]
		\centering
		\begin{tikzpicture}[node distance=1.2cm, auto]
			\node[pnode] (start) {\textbf{70B Enterprise Model}\\High Concurrent Load};
			\node[pnode] (vllm) [below=of start] {\textbf{vLLM (PagedAttention)}\\Reduce KV cache fragmentation; enable continuous batching};
			\node[pnode] (flash) [below=of vllm] {\textbf{FlashAttention-2}\\Reduce HBM access; accelerate prefill};
			\node[pnode] (eagle) [below=of flash] {\textbf{Eagle Speculative Decoding}\\Accelerate autoregressive decode};
			\node[pnode] (end) [below=of eagle] {\textbf{Higher-Throughput Serving System}\\Gains are hardware- and workload-dependent};
			
			\draw[parr] (start) -- node[right] {\small Enable continuous batching} (vllm);
			\draw[parr] (vllm) -- node[right] {\small Accelerate prefilling} (flash);
			\draw[parr] (flash) -- node[right] {\small Accelerate decoding} (eagle);
			\draw[parr] (eagle) -- (end);
		\end{tikzpicture}
		\caption{Optimization Pipeline for Enterprise LLM Serving. Reducing KV cache fragmentation enables continuous batching; prefill and decode are then accelerated independently. Absolute throughput and latency figures depend on hardware configuration, parallelism strategy, and workload distribution.}
		\label{fig:usecase_enterprise}
	\end{figure}
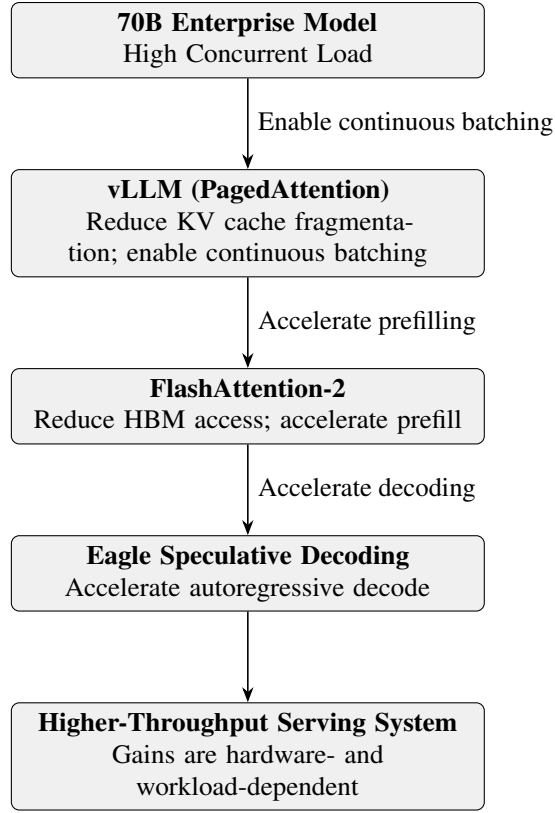

	Bob's binding constraints are throughput and latency. His cluster of A100 GPUs has enough raw VRAM to hold the model parameters, but the dynamic KV cache for many simultaneous sessions causes memory fragmentation and out-of-memory errors under continuous batching. Bob addresses the runtime architecture in three stages. First, he deploys the vLLM serving engine, which uses PagedAttention to manage the KV cache in non-contiguous paged blocks, substantially reducing fragmentation and enabling continuous batching. Second, he enables FlashAttention-2 to reduce HBM traffic during the prefill phase. Third, he integrates Eagle speculative decoding to accelerate the autoregressive decode phase for individual requests. The combined effect on concurrency and latency is configuration-dependent. Actual throughput gains require profiling against the specific GPU count, parallelism strategy, input/output length distribution, and quantization level used in production. Latency targets (whether time-to-first-token, inter-token latency, or end-to-end response time) should also be defined and measured separately, as each optimization affects them differently, as illustrated in Figure~\ref{fig:usecase_enterprise}.

	\subsection{Retrieval-Augmented Generation: Charlie's Legal QA}
	Charlie is developing a legal document QA system. When a user asks a question, the system retrieves 30 pages of dense legal text, resulting in a 64K token context window.

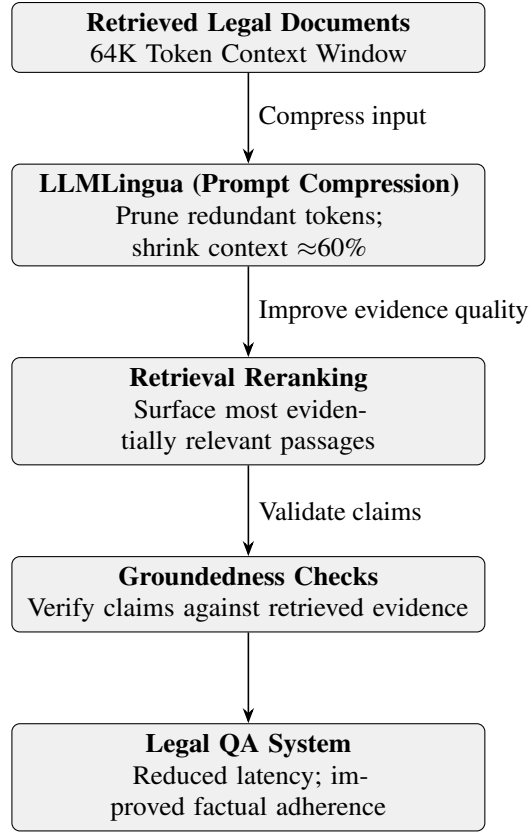
\begin{figure}[h!]
		\centering
		\begin{tikzpicture}[node distance=1.2cm, auto]
			\node[pnode] (start) {\textbf{Retrieved Legal Documents}\\64K Token Context Window};
			\node[pnode] (lingua) [below=of start] {\textbf{LLMLingua (Prompt Compression)}\\Prune redundant tokens; shrink context $\approx$60\%};
			\node[pnode] (rerank) [below=of lingua] {\textbf{Retrieval Reranking}\\Surface most evidentially relevant passages};
			\node[pnode] (ground) [below=of rerank] {\textbf{Groundedness Checks}\\Verify claims against retrieved evidence};
			\node[pnode] (end) [below=of ground] {\textbf{Legal QA System}\\Reduced latency; improved factual adherence};
			
			\draw[parr] (start) -- node[right] {\small Compress input} (lingua);
			\draw[parr] (lingua) -- node[right] {\small Improve evidence quality} (rerank);
			\draw[parr] (rerank) -- node[right] {\small Validate claims} (ground);
			\draw[parr] (ground) -- (end);
		\end{tikzpicture}
		\caption{Optimization Pipeline for Retrieval-Augmented Generation. Prompt compression reduces prefill cost; retrieval reranking improves evidence quality; groundedness checks validate factual adherence. Self-Consistency sampling can supplement this for reasoning stability, but does not solve factuality.}
		\label{fig:usecase_rag}
	\end{figure}

	Charlie faces a latency and accuracy challenge. Processing 64K tokens per query is expensive, and the model occasionally produces unsupported claims about legal provisions. To reduce the prefill cost, Charlie applies LLMLingua: this prompt compression technique evaluates the retrieved documents and prunes redundant tokens before they reach the main LLM, cutting the context size by approximately 60\% while retaining semantically critical content. For the accuracy challenge (hallucinated legal provisions), Charlie implements retrieval reranking to surface the most evidentially relevant passages, paired with post-hoc groundedness checks that verify each generated claim against the retrieved text. While Charlie might also implement Self-Consistency sampling to improve answer stability on complex reasoning queries, he recognizes that Self-Consistency is not a factuality intervention, as correlated samples from the same model will consistently repeat an unsupported claim if the model is confidently wrong. StreamingLLM is not appropriate as the primary long-context strategy here, because it evicts intermediate KV states and cannot provide faithful access to all 64K tokens. It would only be suitable if Charlie were building a stateful dialogue system that streams over a very long interaction history (see Figure~\ref{fig:usecase_rag}).

	\subsection{Production Cost Reduction: Diana's API Scale Problem}
	Diana is a product manager for a consumer AI app that relies entirely on premium, proprietary LLM APIs. Due to a surge in popularity, her monthly API bill has skyrocketed to \$50,000, creating a severe Cost constraint.

\begin{figure}[h!]
		\centering
		\begin{tikzpicture}[node distance=1.2cm, auto]
			\node[pnode] (start) {\textbf{Incoming User Query}};
			\node[pnode] (router) [below=of start] {\textbf{Complexity Classifier}\\Lightweight routing model};
			\node[pnode, text width=4.5cm] (local) [below left=1cm and -1.5cm of router] {\textbf{Local 8B OmniQuant Model}\\Handles low-complexity queries\\Cost: GPU/energy/ops overhead};
			\node[pnode, text width=4.5cm] (premium) [below right=1cm and -1.5cm of router] {\textbf{Premium LLM API}\\Handles high-complexity queries\\Cost: Per-token API pricing};
			\node[pnode] (end) [below=3.5cm of router] {\textbf{Final Response}\\Routing split and savings are traffic-dependent};
			
			\draw[parr] (start) -- (router);
			\draw[parr] (router) -- node[above left, inner sep=1pt] {\small Low Complexity} (local);
			\draw[parr] (router) -- node[above right, inner sep=1pt] {\small High Complexity} (premium);
			\draw[parr] (local) -- (end);
			\draw[parr] (premium) -- (end);
		\end{tikzpicture}
		\caption{Optimization Pipeline for Cascade API Routing. A learned router directs queries to an appropriate model tier. The routing split, cost savings, and quality trade-offs are all traffic-dependent and must be measured on the organization's own workload.}
		\label{fig:usecase_cost}
	\end{figure}
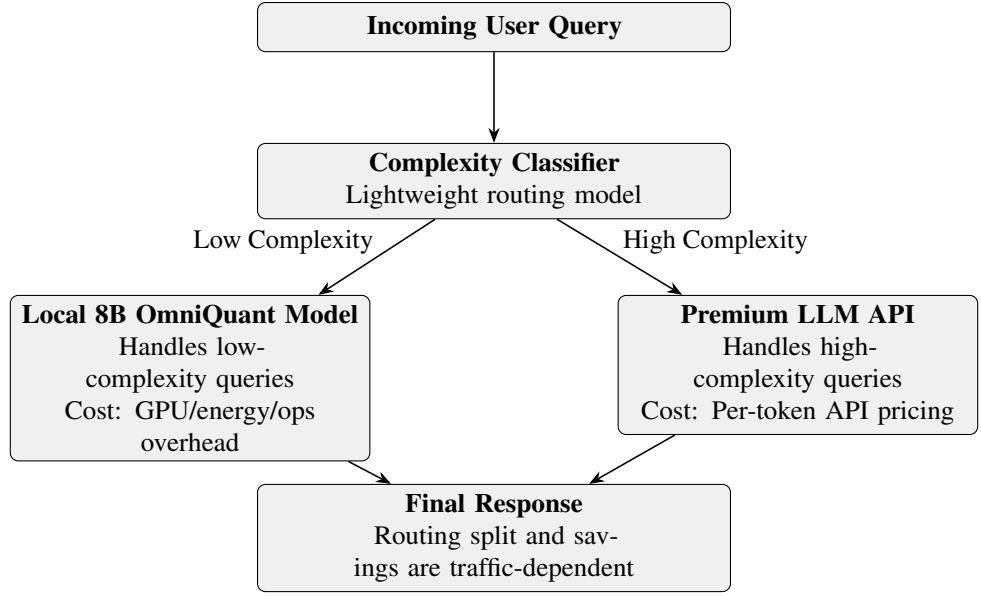

	Diana recognizes that not every query requires the full reasoning capability of the premium API. She designs a FrugalGPT-style cascade router: she deploys a smaller 8B open-weight model on a local cloud instance, quantized with OmniQuant to run efficiently on lower-tier GPUs, and trains a lightweight classifier on her historical traffic to predict query complexity. Simple queries are handled by the local model, and complex queries are escalated to the premium API. The routing split and the resulting cost reduction depend entirely on Diana's own traffic distribution and the quality threshold she sets. The ratios from the original FrugalGPT study are workload-specific and cannot be assumed to transfer directly. The local model also carries real serving costs (GPU or CPU compute, energy, orchestration, and engineering overhead), so Diana must account for these when computing net savings. Diana should also define an accuracy floor per query category and monitor escalation rates continuously, adjusting the router threshold as her traffic mix evolves (Figure~\ref{fig:usecase_cost}).

	\newpage
	\section{Conclusion}
	
	Model optimization is no longer just an algorithmic research pursuit. It has become a foundational engineering discipline required to deploy machine learning economically. By defining the operational space along five interacting constraint dimensions (Data, Latency, Memory, Accuracy, and Retraining) and mapping empirical results from the primary literature to these constraints, practitioners can move from heuristics toward a more structured, evidence-informed methodology. The techniques discussed, ranging from 4-bit quantization and low-rank adaptation to speculative decoding and continuous batching, may be composable when applied to distinct bottlenecks, but their combined effects are not guaranteed to be multiplicative. Interactions can be non-additive, and some combinations introduce quality or latency regressions (for example, quantization reducing speculative-draft acceptance rates, or continuous batching increasing tail latency for individual requests). Practitioners are encouraged to benchmark technique combinations on representative production traffic rather than summing gains reported in independent studies.
	
	\newpage
	\bibliographystyle{plainnat}
	\bibliography{references}

	\newpage
	\appendix
	\section{Technique Comparison}
	
	Table \ref{tab:comparison} provides a summarized comparison of the optimization techniques discussed in this framework, categorizing them by their primary target constraint and key benefits.
	
	\begin{table}[h!]
		\centering
		\caption{Comparison of All Evaluated Model Optimization Techniques}
		\label{tab:comparison}
		\resizebox{\textwidth}{!}{
		\begin{tabular}{llp{9cm}}
			\toprule
			\textbf{Technique} & \textbf{Primary Constraint} & \textbf{Key Benefit / Mechanism} \\
			\midrule
			GPTQ / AWQ & Memory & 3-4 bit post-training quantization with benchmark-dependent accuracy impact. \\
			OmniQuant / ZeroQuant-V2 & Memory & Quantization techniques targeting significant activation outliers. \\
			KIVI / KVQuant & Memory & Tuning-free KV-cache quantization for long-context inference. \\
			Wanda & Memory & Training-free unstructured pruning (weight magnitude $\times$ activation norm); memory/latency benefit requires sparse runtime support. \\
			FlexGen / PowerInfer & Memory & System-level offloading of weights/KV cache to CPU memory and disk. \\
			LoSparse & Memory & Low-rank approximation combined with weight pruning; runtime benefit requires compatible sparse kernels. \\
			FlashAttention / FlashAttention-2 & Latency & IO-aware exact attention; reduces auxiliary attention memory from $O(N^2)$ to $O(N)$ (compute remains $O(N^2)$); speedup is workload- and hardware-dependent. \\
			Medusa / Eagle & Latency & Draft-free speculative decoding for autoregressive speedup. \\
			vLLM (PagedAttention) & Latency & Non-contiguous KV cache management for high-throughput batching. \\
			LLMLingua & Latency & Prompt compression via coarse-to-fine pruning of redundant tokens. \\
			Skeleton-of-Thought (SoT) & Latency & Output organization framework using parallelized batch inference. \\
			StreamingLLM & Latency & Attention sink preservation for stable streaming generation; not suited for full-document recall tasks. \\
			LIMA & Data & High-quality alignment tuning using highly limited (1k) examples. \\
			Socratic CoT & Data & Distillation of synthetic rationales to transfer Chain-of-Thought reasoning. \\
			Gisting / AutoCompressors & Latency & Soft prompt compression mapping task instructions into continuous tokens. \\
			LLM-QAT / Norm Tweaking & Retraining & Original-training-data-free QAT using synthetic data generated from the model itself. \\
			LoRA & Retraining & Low-rank adaptation reducing trainable parameters by up to 10,000x. \\
			LoftQ / QA-LoRA & Retraining & Accuracy-preserving PEFT methods bridged directly with post-training quantization. \\
			Switch Transformers / Expert Choice & Retraining & Mixture-of-Experts routing for scaling capacity with sub-linear compute. \\
			LoraPrune / ZipLM & Retraining & Structured pruning methods integrated dynamically into PEFT pathways. \\
			FrugalGPT & Accuracy (Serving Cost) & Cascade routing between cheap local models and premium APIs. \\
			SpQR / SqueezeLLM & Accuracy & Outlier-aware quantization to protect critical weights in high precision. \\
			CALM / SkipDecode & Accuracy (Serving Cost) & Dynamic early exiting using confidence classifiers across batch inference. \\
			\bottomrule
		\end{tabular}
		}
	\end{table}

\end{document}